\documentclass{article}

\usepackage{svg}
\usepackage[final]{neurips_2024}

\usepackage{siunitx} 
\usepackage{xcolor} 
\usepackage{makecell} 

\usepackage{amsmath,amsfonts,bm}

\def\eqref#1{equation~\ref{#1}}

\def\1{\bm{1}}

\DeclareMathAlphabet{\mathsfit}{\encodingdefault}{\sfdefault}{m}{sl}
\SetMathAlphabet{\mathsfit}{bold}{\encodingdefault}{\sfdefault}{bx}{n}

\usepackage{hyperref}
\usepackage{url}
\usepackage{graphicx} 
\usepackage{epstopdf}
\usepackage{arydshln}
\usepackage{multirow}
\usepackage{wrapfig}
\usepackage{appendix}
\usepackage{booktabs}

\usepackage[utf8]{inputenc} 
\usepackage[T1]{fontenc}    
\usepackage{hyperref}       
\usepackage{url}            
\usepackage{amsfonts}       
\usepackage{nicefrac}       
\usepackage{microtype}      
\usepackage{xcolor}         

\newcommand{\method}{SGEC}

\title{Variational Graph Contrastive Learning}

\author{%
  Shifeng Xie\\
  LTCI, Télécom Paris \\
  Institut Polytechnique de Paris \\
  \texttt{shifeng.xie@telecom-paris.fr} \\
  \And
  Jhony H. Giraldo \\
  LTCI, Télécom Paris \\
  Institut Polytechnique de Paris \\
  \texttt{jhony.giraldo@telecom-paris.fr} \\
}

\begin{document}

\maketitle

\begin{abstract}
Graph representation learning (GRL) is a fundamental task in machine learning, aiming to encode high-dimensional graph-structured data into low-dimensional vectors.
Self-supervised learning (SSL) methods are widely used in GRL because they can avoid expensive human annotation.
In this work, we propose a novel Subgraph Gaussian Embedding Contrast (\method) method.
Our approach introduces a subgraph Gaussian embedding module, which adaptively maps subgraphs to a structured Gaussian space, ensuring the preservation of graph characteristics while controlling the distribution of generated subgraphs.
We employ optimal transport distances, including Wasserstein and Gromov-Wasserstein distances, to effectively measure the similarity between subgraphs, enhancing the robustness of the contrastive learning process.
Extensive experiments across multiple benchmarks demonstrate that \method~outperforms or presents competitive performance against state-of-the-art approaches.
Our findings provide insights into the design of SSL methods for GRL, emphasizing the importance of the distribution of the generated contrastive pairs\footnote{Our code is provided at \url{https://github.com/ShifengXIE/SGEC}}.
\end{abstract}

\section{Introduction}

Graph representation learning (GRL) aims to effectively encode high-dimensional sparse graph-structured data into low-dimensional dense vectors, which is a fundamental task that has been widely studied in a range of fields, including machine learning and data mining \citep{Ju_2024}.

\begin{figure}[t]
\centering
\includegraphics[width=\textwidth]{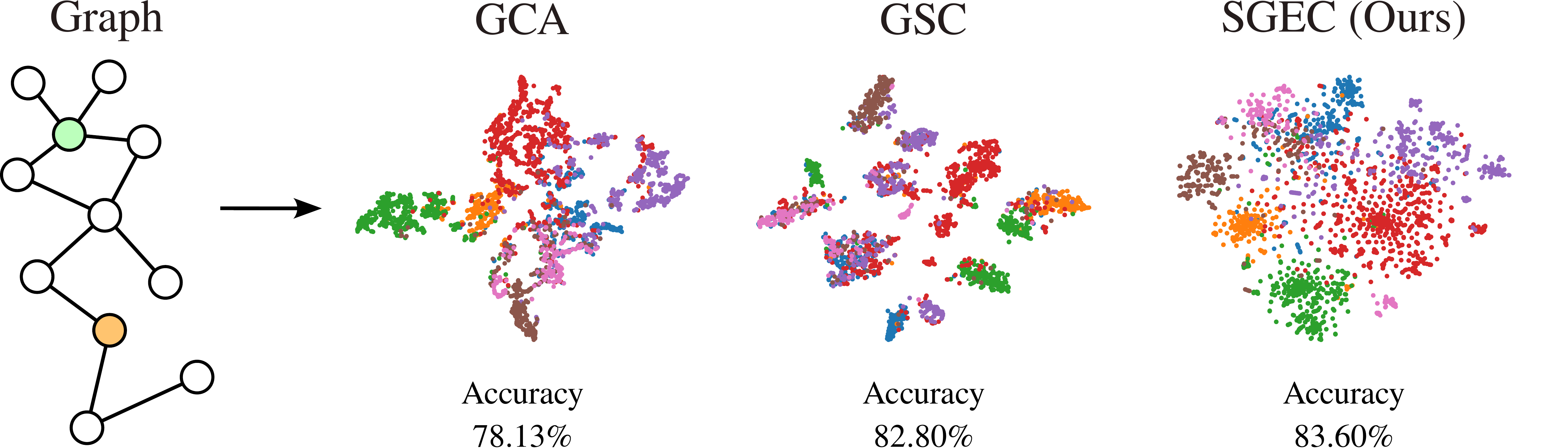}
\caption{T-SNE visualizations of the graph contrastive learning method GCA \citep{GCA}, GSC \citep{GSC}, and our method \method~on the Cora dataset \citep{Cora}.
Each point in the visualization corresponds to a node embedding, with colors indicating classes. \method~maps the node representations into a dense, uniform, and more linearly separable space.
}
\label{fig:graph_representation_comparison}
\end{figure}

Self-supervised learning (SSL) offers a promising approach to GRL by mitigating the need for extensive human annotation \citep{jaiswal2020survey}. 
Particularly, contrastive learning is a prominent approach in SSL that leverages the similarities and differences between data samples or data embeddings to learn representations.
In this context, positive sample pairs are typically two augmented views of the same data point that should be close in the representation space, while negative sample pairs are original and augmented views of different data samples \citep{chen2020simple}.
Current graph-based contrastive learning methods primarily generate positive and negative sample pairs through perturbations \citep{GRACE}\footnote{A complete review of the related work can be found in Appendix \ref{Relative}.}.
However, t-SNE visualizations of current graph-based contrastive learning methods, like GCA \citep{GCA} and GSC \citep{GSC}, reveal uneven node distributions within the same graph, with sharp boundaries and erroneous node clusters as illustrated in Figure~\ref{fig:graph_representation_comparison}.
We observe that those characteristics of previous models penalize their performance in GRL tasks.

In this paper, we propose the Subgraph Gaussian Embedding Contrast (\method) model.
In our method, a subgraph Gaussian embedding (SGE) module is proposed to generate the features of the subgraphs.
The SGE maps input subgraphs to a structured Gaussian space, where the features of the output subgraphs tend towards a Gaussian distribution by using the Kullback–Leibler (KL) divergence. 
SGE is a learnable mapping that controls the distribution of the embeddings.
The embedded subgraphs, paired with the original subgraphs to form positive and negative pairs, then conduct contrastive learning with optimal transport measurements to obtain the graph representation. 
The embedded subgraphs offer diversity to prevent mode collapse (also called positive collapse \citep{jing2022understandingdimensionalcollapsecontrastive}) where the embeddings shrink into a low-dimensional subspace, by controlling the embedding distribution.

For our contributions, firstly, we propose the Subgraph Gaussian Embedding Contrast (\method) model that demonstrates its advantages across eight distinct benchmarks.
We also provide insights into the importance of the distribution of subgraphs generated for contrastive positive and negative pairs.
Finally, we provide different validation and ablation studies to assess the effectiveness of each design choice in \method.

\section{Subgraph Gaussian Embedding Contrast (\method) Method}

\textbf{Mathematical Notation.}
Consider an undirected graph \( G = (\mathcal{V}, \mathcal{E}) \) with vertex set \( \mathcal{V} \) and edge set \( \mathcal{E} \). The feature matrix \( \mathbf{X} = [\mathbf{x}_1, \dots, \mathbf{x}_N]^\top \in \mathbb{R}^{N \times C} \) contains node features \( \mathbf{x}_i \in \mathbb{R}^C \), where \( N \) is the number of nodes and \( C \) is the feature dimension.
The adjacency matrix \( \mathbf{A} \in \mathbb{R}^{N \times N} \) represents the graph topology, and \( \mathbf{D} \) is the diagonal degree matrix.
For some node \( i \), let \( H^i = (\mathcal{V}^i, \mathcal{E}^i) \) be its induced breadth-first search (BFS) subgraph with \( k \) nodes, adjacency matrix \( \mathbf{A}^i \in \mathbb{R}^{k \times k} \), and feature matrix \( \mathbf{X}^i \in \mathbb{R}^{k \times C} \).
Our method embeds this subgraph into a structured space, yielding the embedded subgraph \( \tilde{H}^i = (\mathcal{V}^i, \tilde{\mathcal{E}}^i) \) with adjacency matrix \( \mathbf{A}^i \) and feature matrix \( \tilde{\mathbf{X}}^i \).
We provide the definitions of KL divergence, Wasserstein, and Gromov-Wasserstein distances in Appendix \ref{Preliminaries}.

\begin{figure}[t]
\centering
\includegraphics[width=\textwidth]{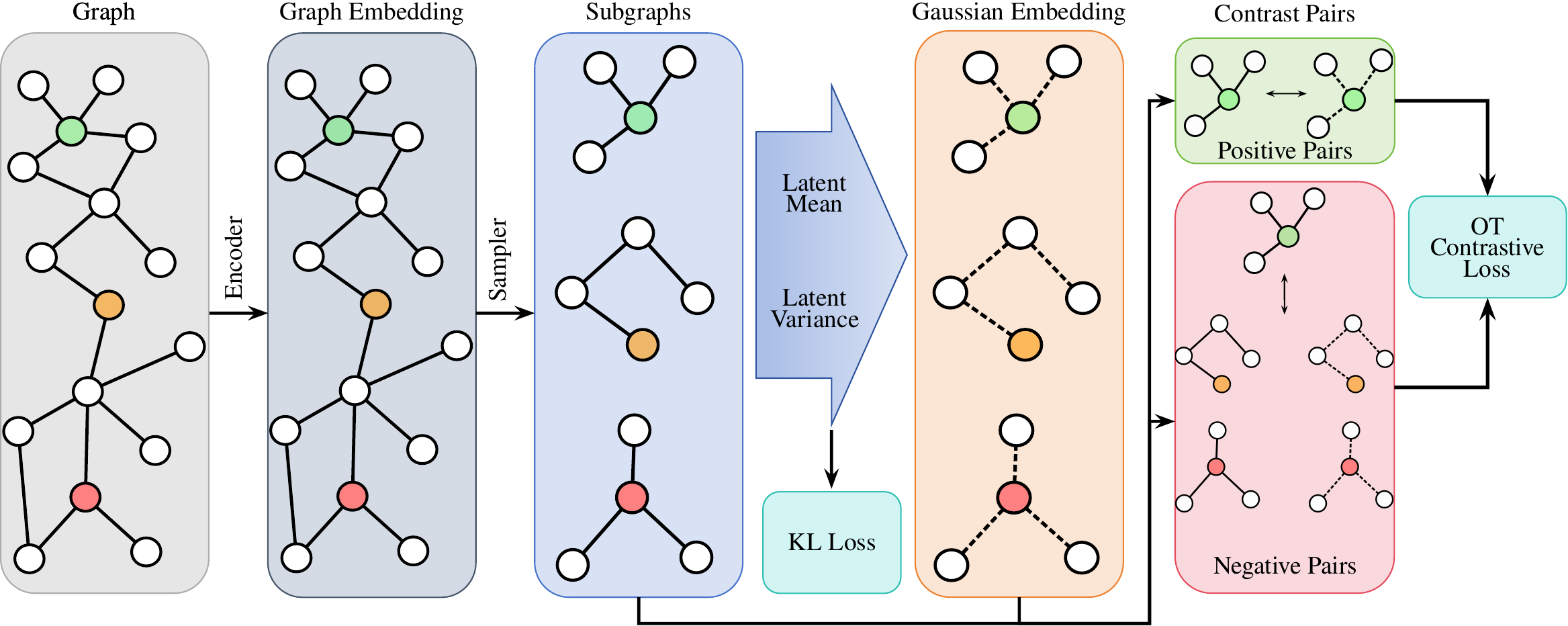}
\caption{Overview of the Subgraph Gaussian Embedding Contrast (\method) method. The SGEC method begins with a graph encoder to generate embeddings and employs a breadth-first search to sample subgraphs for some set of nodes $\mathcal{S}$.
A Gaussian embedding module then produces contrastive samples in a Gaussian space of these subgraphs.
Positive pairs consist of subgraphs with the same central node, while negative pairs have different central nodes. SGEC introduces the Wasserstein and Gromov-Wasserstein distances to compute distances between subgraphs for contrastive learning.}
\label{fig:method_overview}
\end{figure}

\textbf{Overview of the Proposed Method.}
Figure~\ref{fig:method_overview} shows an overview of our methodology, where our process begins with an encoder of the input graph.
Subsequently, we obtain subgraphs utilizing BFS-based sampling.
The node representations and topology information within these subgraphs are adaptively embedded into a latent space tending towards a Gaussian distribution, which is enforced by the KL divergence regularization.
Finally, we use the Wasserstein and the Gromov-Wasserstein distance to measure dissimilarities in the subgraphs for contrastive learning.

\textbf{Graph Encoder.}
We begin by employing a graph encoder to preprocess the graph data \citep{GAE, GSC}.
The graph encoder comprises some graph convolutions layers and activation functions $\sigma(\cdot)$.
We use two graph convolution layers \citep{kipf2016semi} in \method.
Further details on the implementation of these layers are provided in the Appendix \ref{Encoder}.

\textbf{Subgraph Gaussian Embedding (SGE).}
The SGE module comprises GraphSAGE \citep{liu2020graphsage} and GAT (graph attention) \citep{velivckovic2017graph} layers.
Appendix \ref{GraphSAGE} provides further details about GraphSAGE.
We first randomly sample a set of nodes $\mathcal{S}$ and get their BFS-induced subgraphs, where $\vert \mathcal{S}\vert$ is a hyperparameter of \method.
We then apply the combination of GraphSAGE and GAT layers on these subgraphs.
The latent means and variances are encoded by separate GAT networks as follows:
\[
\boldsymbol{\mu}_i = \text{GATConv}_\mu(\mathbf{h}_i), \quad \log \boldsymbol{\sigma}_i^2 = \text{GATConv}_\sigma(\mathbf{h}_i), \quad \mathbf{\tilde{x}}_i = \boldsymbol{\mu}_i + \exp(\log(\boldsymbol{\sigma_i})) \odot \boldsymbol{\epsilon},
\]
where \(\boldsymbol{\epsilon} \sim \mathcal{N}(\mathbf{0}, \mathbf{I})\) represents Gaussian noise drawn from the standard normal distribution, \(\mathbf{h}_i\) represents the output of the last GraphSAGE layer, \(\mathbf{\tilde{x}}_i\) denote the embedded features of node $i$. 
Note that $i$ is some node in the set of $\vert \mathcal{S}\vert$ induced subgraphs.
In this configuration, $\boldsymbol{\mu}_i$ and $\log \boldsymbol{\sigma}_i^2$ are derived through separate GAT layers.
SGE effectively capture both the node attributes and the graph topology. In our method, we preserve the original graph topology, so the embedded subgraphs have the same adjacency matrices as the original subgraphs. 

\textbf{Regularization in the Subgraph Gaussian Embedding.}
In our approach, we introduce a regularization constraint to the SGE module to guide the embedded subgraph node features towards a Gaussian distribution.
For the motivation of introducing the Gaussian regularization, please see the Appendix \ref{Motivation}.
This regularization is implemented using the Kullback-Leibler divergence, denoted as \( \text{KL}(q(\cdot) \| p(\cdot)) \), between the embedding distribution \( q(\cdot) \) and a predefined Gaussian prior. 
The prior \( p(\tilde{\mathbf{X}}) \) is taken as a product of independent Gaussian distributions for each latent variable \( \tilde{\mathbf{x}}_i \), given by \( p(\mathbf{\tilde{x}}_i) = \mathcal{N}(\mathbf{\tilde{x}}_i | \mathbf{0}, \mathbf{I}) \). 
The expression for the KL divergence then simplifies to:
\[
\text{KL}\left(q(\tilde{\mathbf{X}} \mid \mathbf{X}, \mathbf{A}) \,\|\, p(\tilde{\mathbf{X}})\right) = \frac{1}{\vert \mathcal{P} \vert} \sum_{i \in \mathcal{P} } \sum_{j=1}^{\tilde{C}} \left( \boldsymbol{\mu}_{ij}^2 + \boldsymbol{\sigma}_{ij}^2 - 1 - 2 \log \boldsymbol{\sigma}_{ij} \right),
\]
where \( \boldsymbol{\mu} \) and \( \boldsymbol{\sigma}^2 \) are the latent mean and latent variance of the $i$th embedded node, respectively, and $\mathcal{P}$ is the set of nodes in the $\vert \mathcal{S} \vert$ induced subgraphs.
\(\tilde{C}\) is the dimension of the latent features.

\textbf{Optimal Transport Contrastive Loss.}
Our contrastive loss function integrates the Wasserstein and Gromov-Wasserstein distances into the InfoNCE loss formulation \citep{oord2019representationlearningcontrastivepredictive}, addressing the complexities of graph-based data. 
The Wasserstein distance, \(W(\mathbf{X}^i, \tilde{\mathbf{X}}^i)\), captures feature distribution representation within subgraphs, whereas the Gromov-Wasserstein distance, \(GW(\mathbf{A}^i, \mathbf{X}^i, \mathbf{A}^i, \tilde{\mathbf{X}}^i)\), captures structural discrepancies, providing a topology-aware similarity measure.
The complete contrastive loss \(\mathcal{L}_{\text{contrast}}\) is the sum of \(\mathcal{L}_{\text{W}}\) and \(\mathcal{L}_{\text{GW}}\) provided as follows:
\begin{equation}
\nonumber
\begin{split}
&\resizebox{0.86\textwidth}{!}{$\mathcal{L}_{\text{W}} = \alpha \left(- \sum_{i \in \mathcal{S}} \log \frac{\exp(-W(\mathbf{X}^i, \tilde{\mathbf{X}}^i)/\tau)}{\sum_{j \in \mathcal{S}, j\neq i}^{N} \left(\exp(-W(\mathbf{X}^i, \tilde{\mathbf{X}}^j)/\tau)+ \exp(-W(\mathbf{X}^i, \mathbf{X}^j)/\tau)\right))} \right),$}\\
&\resizebox{\textwidth}{!}{$\mathcal{L}_{\text{GW}} = (1 - \alpha) \left(- \sum_{i \in \mathcal{S}} \log \frac{\exp(-GW(\mathbf{A}^i,\mathbf{X}^i, \mathbf{A}^i, \tilde{\mathbf{X}}^i)/\tau)}{\sum_{j\in \mathcal{S}, j\neq i}^{N} \left(\exp(-GW(\mathbf{A}^i,\mathbf{X}^i, \mathbf{A}^j, \tilde{\mathbf{X}}^j)/\tau)+ \exp(-GW(\mathbf{A}^i,\mathbf{X}^i, \mathbf{A}^j,\mathbf{X}^j)/\tau)\right)}\right),$}
\end{split}
\end{equation}
where \(\alpha\) is a hyperparameter that balances the emphasis on feature distribution and structural fidelity, and $\tau$ is a temperature hyperparameter.
The final loss \( \mathcal{L} \) of our model incorporates both contrastive and regularization components balanced by a hyperparameter \( \beta \) as follows:
\(
\mathcal{L} = \mathcal{L}_{\text{contrast}} + \beta \text{KL}(q(\mathbf{\tilde{X}} \mid \mathbf{X}, \mathbf{A}) \| p(\mathbf{\tilde{X}})).
\)

\section{Experimental Evaluation}

\textbf{Comparison of Classification Performance.}
We compare our model against state-of-the-art SSL for node classification methods: GREET \citep{GREET}, GRACE \citep{GRACE}, GSC \citep{GSC}, and MUSE \citep{MUSE}. Additionally, we include classic SSL baselines for a comprehensive comparison: DGI \citep{DGI}, GCA \citep{GCA}, and GraphMAE \citep{graphmae}.
We test \method~on standard benchmark datasets for node classification Cora, Citeseer, Pubmed, Coauthor, Squirrel, Chameleon, Cornell, and Texas \citep{giraldo2023trade}.
Please see the Appendix \ref{Experimental} for more information about the datasets and experimental setup.
The evaluation results for these eight datasets are summarized in Table \ref{tab:performance}.
We observe that SGEC achieves the highest accuracies on the Squirrel, Cornell, and Texas datasets, outperforming existing state-of-the-art SSL methods.
While not always leading on other datasets, SGEC consistently demonstrates competitive performance, highlighting its robustness and effectiveness in GRL.

\textbf{Validation and Ablation Studies.}
We design four network configurations for validation and ablation studies of \method~in Table \ref{tab:validation}.
The first, termed ``W.O. Regularization", omits the KL divergence component, retaining the original network structure but mapping node features into an unstructured space.
This ablation demonstrates the efficacy of the added regularization term encouraging a Gaussian distribution.
The second study, ``with decoder", incorporates a decoder of two fully connected layers. 
This allows the contrastive learning loss to substitute for the reconstruction loss and effectively models the network as a Variational Autoencoder (VAE) generating contrastive pairs.
This study illustrates that \method~is not merely a combination of a VAE model with contrastive learning.
The third study, ``with reconstruction", includes an $\ell_1$ reconstruction loss calculated as the norm of the difference between input and output features, adding a constraint to enforce similarity between them.
The fourth study, ``Dropout", replaces the KL divergence with dropout as the regularization technique.
Our method outperforms dropout, validating its superiority in SSL tasks for GRL.



\begin{table}[t]
\centering
\caption{Comparison (accuracy \%) of \method~against state-of-the-art SSL method for graphs.}
\label{tab:performance}
\resizebox{\linewidth}{!}{
\begin{tabular}{@{}lSSSSSSSS@{}}
\toprule
\textbf{Method} & {\textbf{Cora}} & {\textbf{Citeseer}} & {\textbf{Pubmed}} & {\textbf{Coauthor}} & {\textbf{Squirrel}} & {\textbf{Chameleon}} & {\textbf{Cornell}} & {\textbf{Texas}} \\
\midrule
MUSE      & {69.90$_{\pm 0.41}$} & {66.35$_{\pm 0.40}$} & {79.95$_{\pm 0.59}$} & {90.75$_{\pm 0.39}$} & {40.15$_{\pm 3.04}$} & {55.59$_{\pm 2.21}$} & {83.78$_{\pm 3.42}$} & {83.78$_{\pm 2.79}$} \\
GREET     & \textbf{84.40$_{\pm 0.77}$} & \textbf{74.10$_{\pm 0.44}$} & {80.29$_{\pm 0.24}$} & \textbf{94.65$_{\pm 0.18}$} & {39.76$_{\pm 0.75}$} & {60.57$_{\pm 1.03}$} & {78.36$_{\pm 3.77}$} & {78.03$_{\pm 3.94}$} \\
GRACE     & {83.30$_{\pm 0.74}$} & {72.10$_{\pm 0.60}$} & \textbf{86.70$_{\pm 0.16}$} & {92.78$_{\pm 0.04}$} & {52.10$_{\pm 0.94}$} & {52.29$_{\pm 1.49}$} & {60.66$_{\pm 11.32}$} & {75.74$_{\pm 2.95}$} \\
GSC       & {82.80$_{\pm 0.10}$} & {71.00$_{\pm 0.10}$} & {85.60$_{\pm 0.20}$} & {91.88$_{\pm 0.11}$} & {51.32$_{\pm 0.21}$} & {64.02$_{\pm 0.29}$} & {93.56$_{\pm 1.73}$} & {88.64$_{\pm 1.21}$} \\
DGI       & {81.99$_{\pm 0.95}$} & {71.76$_{\pm 0.80}$} & {77.16$_{\pm 0.24}$} & {92.15$_{\pm 0.63}$} & {38.80$_{\pm 0.76}$} & {58.00$_{\pm 0.70}$} & {70.82$_{\pm 7.21}$} & {81.48$_{\pm 2.79}$} \\
GCA       & {78.13$_{\pm 0.85}$} & {67.81$_{\pm 0.75}$} & {80.63$_{\pm 0.31}$} & {93.10$_{\pm 0.20}$} & {47.13$_{\pm 0.61}$} & {56.54$_{\pm 1.07}$} & {53.11$_{\pm 9.34}$} & {81.02$_{\pm 2.30}$} \\
GraphMAE  & {84.20$_{\pm 0.40}$} & {73.40$_{\pm 0.40}$} & {81.10$_{\pm 0.40}$} & {80.63$_{\pm 0.15}$} & {48.26$_{\pm 1.21}$} & \textbf{71.05$_{\pm 0.36}$} & {61.93$_{\pm 4.59}$} & {67.80$_{\pm 3.37}$} \\ 
\hdashline
SGEC (Ours)            & {83.60$_{\pm 0.10}$} & {73.14$_{\pm 0.14}$} & {84.60$_{\pm 0.10}$} & {92.34$_{\pm 0.04}$} & \textbf{56.39$_{\pm 0.57}$} & {69.14$_{\pm 1.12}$} & \textbf{94.58$_{\pm 2.13}$} & \textbf{92.38$_{\pm 0.81}$} \\
\bottomrule
\end{tabular}
}
\end{table}

\begin{table}[t]
\centering
\caption{Validation and ablation studies on different modules of \method.}
\label{tab:validation}
\resizebox{\linewidth}{!}{
\begin{tabular}{@{}lcccccccc@{}}
\toprule
\textbf{Methodology} & \textbf{Cora} & \textbf{Citeseer} & \textbf{Pubmed} & \textbf{Coauthor} & \textbf{Squirrel} & \textbf{Chameleon} & \textbf{Cornell} & \textbf{Texas} \\
\midrule
W.O. Regularization     & 83.00$_{\pm 0.07}$ & 71.88$_{\pm 0.07}$ & \textbf{85.46$_{\pm 0.04}$} & 91.96$_{\pm 0.09}$ & 42.99$_{\pm 0.17}$ & 64.25$_{\pm 0.21}$ & \textbf{94.58$_{\pm 0.22}$} & 75.72$_{\pm 0.40}$ \\
With Decoder  & 82.80$_{\pm 0.07}$ & 73.00$_{\pm 0.08}$ & 80.26$_{\pm 0.37}$ & 92.03$_{\pm 0.13}$ & 35.47$_{\pm 0.21}$ & 60.30$_{\pm 0.21}$ & 93.56$_{\pm 0.64}$ & 88.98$_{\pm 0.52}$ \\
With reconstruction   & 81.60$_{\pm 0.99}$ & 69.60$_{\pm 0.10}$ & 67.54$_{\pm 0.35}$ & 87.03$_{\pm 0.65}$     & 30.52$_{\pm 0.48}$ & 43.26$_{\pm 0.66}$ & 53.29$_{\pm 0.13}$ & 63.45$_{\pm 0.48}$ \\
Dropout   & 79.00$_{\pm 0.21}$ & 70.60$_{\pm 3.52}$ & 80.84$_{\pm 0.05}$ & 91.52$_{\pm 0.37}$     & 44.24$_{\pm 0.50}$ & 58.94$_{\pm 0.72}$ & 85.76$_{\pm 0.24}$ & 87.98$_{\pm 0.15}$ \\
\hdashline
SGEC (Ours) & \textbf{83.60$_{\pm 0.10}$} & \textbf{73.14$_{\pm 0.14}$} & {84.60$_{\pm 0.10}$} & \textbf{92.34$_{\pm 0.04}$} & \textbf{56.39$_{\pm 0.57}$} & \textbf{69.14$_{\pm 1.12}$} & 94.57$_{\pm 2.13}$ & \textbf{92.38$_{\pm 0.81}$} \\
\bottomrule
\end{tabular}
}
\end{table}

\section{Conclusion}

We introduced \method, which embeds subgraphs into a Gaussian space to enhance self-supervised graph representation learning.
By controlling embedding distributions and utilizing optimal transport distances, \method~achieves superior or competitive performance.
Our results emphasize the importance of distribution control in generating contrastive pairs.

Future work will involve integrating spectral-based contrastive learning methods \citep{polygcl} to further validate and enhance our proposed approach.
Additionally, we intend to explore the applicability of our framework to other modalities, extending beyond graph data. 
This would involve validating our underlying ideas in different contexts, potentially broadening the impact and utility of our approach in various domains of representation learning.

\bibliographystyle{unsrtnat}
\bibliography{neurips_2024}

\newpage

\appendix

\section*{Appendix}


In the appendix, we provide supplementary material that includes a concise review of related work on graph representation learning and self-supervised learning on graphs.
We also present essential mathematical preliminaries for our approach and discuss the motivation for introducing KL regularization into our method.
Finally, we offer detailed experimental evaluations and dataset descriptions, implementation details, and sensitivity analyses of the regularization constraint and subgraph size.

\section{Related Work}
\textbf{Graph Neural Networks.}
\label{Relative}

Graph Neural Networks (GNNs) play a crucial role in learning representations that effectively capture both node features and graph topology \citep{9046288}.
Among various architectures, Graph Convolutional Networks (GCNs), pioneered by \citet{kipf2016semi}, utilize a special approach inspired by spectral graph convolutions.
GraphSAGE, introduced by \citet{hamilton2017inductive}, employs multiple aggregator functions and enables inductive learning on graphs.
\citet{velivckovic2017graph} proposed Graph Attention Networks (GAT), which incorporate attention mechanisms to dynamically weigh the significance of neighboring nodes.

\textbf{Self-Supervised Learning on Graphs.}
Self-supervised learning (SSL) on graphs can be categorized into four main frameworks \citep{liu2022graph}: generation-based \citep{GAE, mgae}, auxiliary property-based \citep{cagnn}, contrastive-based \citep{deepwalk, node2vec}, and hybrid methods \citep{gpt}, each utilizing distinct pretext tasks and objective functions. 
Generation-based methods like Graph Autoencoders (GAEs) \citep{GAE} use node attributes and graph structure for learning representations.
In auxiliary property-based SSL \citep{cagnn}, properties such as node roles are predicted to facilitate learning, while contrastive methods like \textit{Deep Graph Infomax} (DGI) \citep{DGI} maximize mutual information between different graph patches.
Hybrid approaches combine these strategies to enhance model robustness and adaptability \citep{liu2022graph}.

\textbf{Generative Self-supervised Learning on Graphs.}
Graph data requires additional processing compared to standard vector-based data due to the interrelationship of edge and node features. Generative SSL methods often take full graphs or subgraphs as input, aiming to reconstruct either node features or the graph's topology. This splits into feature generation focusing on node or edge features \citep{nodecluster,m3s} and structure generation \citep{pothen,optimal,gailhard2024hygene} targeting the graph's topological aspects, thus deepening the understanding of graph properties crucial for downstream tasks.

\textbf{Augmented Contrastive Learning.}
Innovative methodologies in augmented contrastive learning for graphs have included: 1) graph augmentations that generate diverse graph instances \citep{GraphECL}; 2) various contrastive pretext tasks designed for non-Euclidean spaces \citep{GSC}; and 3) mutual information estimation techniques that underpin learning objectives alongside specific pretext tasks \citep{DGI, GCA}. These approaches leverage the structural nuances of graphs to improve the quality and utility of the learned representations.

\textbf{Positioning of \method.}
Our approach integrates subgraph feature generation with optimal transport measurement to perform subgraph-wise contrastive learning. 
In our approach, the features of the sampled subgraphs are generated by SGE, compared with generative SSL methods, we adopt contrastive learning as the training paradigm.
Furthermore, we use the Wasserstein and Gromov-Wasserstein distances to measure the distance between subgraphs.
To the best of our knowledge, the most similar approach to ours is GSC \citep{GSC}, focusing on the distribution of generated features, our method aims to avoid the sparse of generated node vector features, achieving superior graph representations. 
This methodology not only captures the intrinsic properties of nodes and topological information but also maintains a distribution that enhances generalizability and effectiveness in downstream applications.~

\section{Preliminaries}
\label{Preliminaries}
\subsection{Optimal Transport Distance}
The Wasserstein distance \citep{wasserstein}, commonly used in optimal transport theory, serves as a robust metric for comparing probability distributions defined over metric spaces. For subgraphs \( H^i \) and \( H^j \), their corresponding feature matrices are denoted as \( \mathbf{X}^i \in \mathbb{R}^{k^i \times C} \) and \( \mathbf{X}^j \in \mathbb{R}^{k^j \times C} \), where \( k^i \) and \( k^j \) represent the number of nodes in the respective subgraphs, and \( C \) is the feature dimension. The Wasserstein distance between the feature distributions of these subgraphs is defined as:
\[
W(\mathbf{X}^i, \mathbf{X}^j) = \min_{\mathbf{T} \in \pi(\mathbf{u}, \mathbf{v})} \sum_{p=1}^{k^i} \sum_{q=1}^{k^j} T_{pq} c(\mathbf{X}^i_p, \mathbf{X}^j_q),
\]
where \( \mathbf{T} \in \pi(\mathbf{u}, \mathbf{v}) \) represents the set of all possible transport plans with marginals \( \mathbf{u} \) and \( \mathbf{v} \), which are the node feature distributions in subgraphs \( H^i \) and \( H^j \), respectively. \( T_{pq} \) is the transportation plan between nodes \( p \) and \( q \), and \( c(\mathbf{X}^i_p, \mathbf{X}^j_q) \) represents the cost function, often defined as \( \exp\left( -\frac{\langle \mathbf{X}^i_p, \mathbf{X}^j_q \rangle}{\tau} \right) \), where \( \langle \cdot, \cdot \rangle \) denotes the cosine similarity between node features, and \( \tau \) is a temperature parameter. This distance captures the minimal cost of transforming one subgraph's node feature distribution into another.

Similarly, the Gromov-Wasserstein distance \citep{arya2024gromovwassersteindistancespheres} extends this idea to compare graph-structured data, where the internal distances between nodes are taken into account. For two subgraphs \( H^i \) and \( H^j \) with adjacency matrices \( \mathbf{A}^i \in \mathbb{R}^{k^i \times k^i} \) and \( \mathbf{A}^j \in \mathbb{R}^{k^j \times k^j} \), and feature matrices \( \mathbf{X}^i \) and \( \mathbf{X}^j \), the Gromov-Wasserstein distance is defined as:
\[
GW(\mathbf{A}^i, \mathbf{A}^j, \mathbf{X}^i, \mathbf{X}^j) = \min_{\mathbf{T} \in \pi(\mathbf{u}, \mathbf{v})} \sum_{p,\tilde{p},q,\tilde{q}} T_{pq} T_{\tilde{p}\tilde{q}} \left| d_{\mathbf{A}^i}(\mathbf{X}^i_p, \mathbf{X}^i_{\tilde{p}}) - d_{\mathbf{A}^j}(\mathbf{X}^j_q, \mathbf{X}^j_{\tilde{q}}) \right|,
\]
where \( d_{\mathbf{A}^i}(\mathbf{X}^i_p, \mathbf{X}^i_{\tilde{p}}) \) and \( d_{\mathbf{A}^j}(\mathbf{X}^j_q, \mathbf{X}^j_{\tilde{q}}) \) represent the shortest path distances between node pairs \( (p, \tilde{p}) \) in subgraph \( H^i \), and \( (q, \tilde{q}) \) in subgraph \( H^j \), respectively. The matrix \( \mathbf{T} \in \pi(\mathbf{u}, \mathbf{v}) \) is the optimal transport plan that matches node pairs from the two subgraphs. This metric assesses the discrepancy between the internal node pair distances of the two subgraphs, aligning their intrinsic geometric structures.

\subsection{The Kullback-Leibler (KL) Divergence}

The Kullback-Leibler (KL) divergence is an asymmetry measure between two probability distributions \citep{van2014renyi}, \( P \) and \( Q \). It quantifies the amount of information lost when \( Q \) is used to approximate \( P \). In mathematical terms, the KL divergence is defined as:
\[
D_{KL}(P \| Q) = \sum_{x \in \mathcal{X}} P(x) \log \frac{P(x)}{Q(x)},
\]
where \( P(x) \) and \( Q(x) \) are the probability masses of \( P \) and \( Q \) at each point \( x \) in the sample space \( \mathcal{X} \).

\subsection{Graph Convolutional Network}
\label{Encoder}
The graph encoder uses two graph convolution layers, which are mathematically represented as follows:
\[
\mathbf{H}_{1} = \sigma\left((\mathbf{D}^{-\frac{1}{2}} \left(\mathbf{A}+ \mathbf{I}\right) \mathbf{D}^{-\frac{1}{2}}  \mathbf{X} \mathbf{W}_1\right), \quad
\mathbf{H}_2 = \sigma\left(\mathbf{D}^{-\frac{1}{2}} \left(\mathbf{A}+ \mathbf{I}\right) \mathbf{D}^{-\frac{1}{2}} \mathbf{H}_1 \mathbf{W}_2\right).
\]

\subsection{GraphSAGE Layer}
\label{GraphSAGE}

For GraphSAGE, we present expressions:
\[
\mathbf{x}_v^{(l+1)} = \sigma \left( \mathbf{W}^{(l)} \cdot \text{MEAN} \left( \mathbf{x}_v^{(l)}, \{\mathbf{x}_u^{(l)} \mid u \in \mathcal{N}(v)\} \right) + \mathbf{b}^{(l)} \right),
\]
where \(\mathbf{x}_v^{(l)}\) is the feature vector of node \(v\) at layer \(l\), \(\mathbf{W}^{(l)}\) is the weight matrix at layer \(l\), \(\mathcal{N}(v)\) denotes the set of neighbors of node \(v\).
\section{Motivation of Introducing Regularization}
\label{Motivation}
Our method involves forming positive and negative pairs from both the structured feature representations and the original input features for contrastive learning. This pairing strategy offers multiple advantages: it enriches the combination of features, reduces the risk of overfitting, and smooths and linearizes the feature distribution.

We incorporate a Gaussian structure into the embeddings by enforcing the embedded subgraph node features to approximate a Gaussian distribution. This is regulated using the Kullback-Leibler (KL) divergence from the distribution \(q(\mathbf{\tilde{X}} \mid \mathbf{X}, \mathbf{A})\) of the embeddings to a standard Gaussian prior \(p(\mathbf{\tilde{x}})\), given by:
\[
-\log p(\mathbf{\tilde{x}} \mid \mathbf{x}, \mathbf{A}) = \frac{(\mathbf{x} - \mathbf{\tilde{x}})^2}{2\boldsymbol{\sigma}^2} + \text{const},
\]
where "const" denotes a normalization constant, irrelevant to our optimization process. \( p(\mathbf{\tilde{x}}) \) is assumed to be a Gaussian distribution, and the formula represents the negative log-likelihood. This term is precisely what we need when calculating the KL divergence, as it evaluates the log-likelihood of the observation \( \mathbf{\tilde{x}} \) relative to the assumed distribution \( p(\mathbf{\tilde{x}}) \). This formulation allows us to evaluate the expectation of the KL divergence as follows:

\[
\mathbb{E}_{q(\mathbf{\tilde{X}} \mid \mathbf{X}, \mathbf{A})} \left[ \frac{(\mathbf{x}-\mathbf{\tilde{x}})^2}{2\boldsymbol{\sigma}^2} \right].
\]

To compute this expectation, we assume that the embedding distribution \(q(\mathbf{\tilde{X}} \mid \mathbf{X}, \mathbf{A})\) centers around \(\mathbf{X}\) with variance \(\boldsymbol{\sigma}^2\), reflecting how \(\mathbf{\tilde{X}}\) is encoded from \(\mathbf{X}\). In our Optimal Transport Contrastive Loss section, the Wasserstein distance \(W(\mathbf{X}^i, \mathbf{\tilde{X}}^i)\) plays a crucial role in measuring the cost of transporting one probability distribution to another, enhancing the model's sensitivity to the geometrical structure of the data:
\[
W(\mathbf{X}^i, \mathbf{\tilde{X}}^i) = \min_{\mathbf{T} \in \pi(\mathbf{u}, \mathbf{v})} \sum_{p=1}^{k^i} \sum_{q=1}^{k^i} T_{pq} c(\mathbf{X}^i_p, \mathbf{\tilde{X}}^i_q) = \inf_{\gamma \in \Gamma(\mu_{\mathbf{X}^i}, \mu_{\mathbf{\tilde{X}^i}})} \int \|\mathbf{x} - \mathbf{\tilde{x}}\| d\gamma(\mathbf{x}, \mathbf{\tilde{x}}). 
\]
\(\Gamma(\mu_{\mathbf{X}^i}, \mu_{\mathbf{\tilde{X}}^i})\) denotes the set of all joint distributions that relate the distributions \(\mu_{\mathbf{X}^i}\) and \(\mu_{\mathbf{\tilde{X}}^i}\). These distributions represent the probabilistic distributions of two different random variables, \(\mathbf{X}^i\) and \(\mathbf{\tilde{X}}^i\), respectively. This Wasserstein distance provides a robust measure of the discrepancy between the embedding distribution and the Gaussian prior, effectively guiding the embeddings to conform to the desired Gaussian structure.

\section{Experimental Evaluation}
\label{Experimental}

\textbf{Datasets.} In this study, we have selected several widely used datasets for graph node classification to conduct a comprehensive evaluation of our proposed method. These datasets encompass various types of networks including academic citation networks, collaboration networks, and web page networks, providing a diverse set of challenges and characteristics.
Table \ref{tab:selected_datasets} summarizes the basic statistics of these datasets:

\begin{table}
\centering
\caption{Overview of selected datasets used in the study.}
\label{tab:selected_datasets}
\resizebox{\linewidth}{!}{
\begin{tabular}{@{}lcccccc@{}}
\toprule
Dataset    & \# Nodes & \# Edges   & \# Features & Avg. \# degree & \# Classes \\ 
\midrule
Cora \citep{Cora}       & 2,708     & 5,429       & 1,433        & 4.0            & 7          \\
Citeseer \citep{citeseer}   & 3,312    & 4,732      & 3,703       & 2.9            & 6          \\
Pubmed \citep{Pubmed}     & 19,717   & 44,338    & 500         & 4.5            & 3          \\
Coauthor.CS \citep{coauther}   & 18,333   & 163,788     & 6,805       & 17.9            & 15         \\
Squirrel \citep{Squirrel}   & 5,201    & 217,073    & 2,089       & 83.5           & 5          \\
Chameleon \citep{Chameleon}  & 2,277    & 36,101     & 2,325       & 31.7           & 5          \\
Cornell \citep{Cornell}    & 183      & 298        & 1,703       & 3.3            & 5          \\
Texas \citep{Cornell}      & 183      & 325        & 1,703       & 3.6            & 5          \\
\bottomrule
\end{tabular}
}
\end{table}

\textbf{Implementation Details.}
The model's architecture leverages the strengths of well-established libraries such as \texttt{PyTorch}, \texttt{PyTorch Geometric}, and \texttt{Optuna}. It is optimized for execution on contemporary GPU architectures, including the RTX 3060 and A40, ensuring broad applicability. Our approach adopts a self-supervised scheme evaluated via linear probing. The model is trained using the official training subsets of the referenced datasets. Hyperparameter tuning involves a random search on the validation set to determine optimal values for learning rate, alpha, and beta. The identified best configuration is subsequently employed for tests on the dataset. Loss minimization is performed using the Adam optimizer, which produces a series of node embeddings. These embeddings are crucial for the downstream task of node classification, demonstrating the model's efficacy and adaptability to diverse hardware and data scenarios.

\textbf{Hyperparameter Random Search.}
Informed by the findings of \citet{gasteiger, topping2022understandingoversquashingbottlenecksgraphs, giraldo2023trade}, which indicate the sensitivity of graph neural networks to hyperparameter settings, we undertake an extensive random search to optimize our model. The training proceeds on the official splits of the train datasets, with the random search conducted on the validation dataset to pinpoint the best configurations. These settings are then implemented to evaluate the model on the test dataset, thereby confirming the robustness of our model under optimal hyperparameter settings.
The ranges of the hyperparameters explored and the code implementation of our approach will be made publicly available to facilitate replication and further research.

\subsection{Sensitivity Analysis of the Regularization Constraint}
\begin{figure}
\centering
\includegraphics[width=0.7\textwidth]{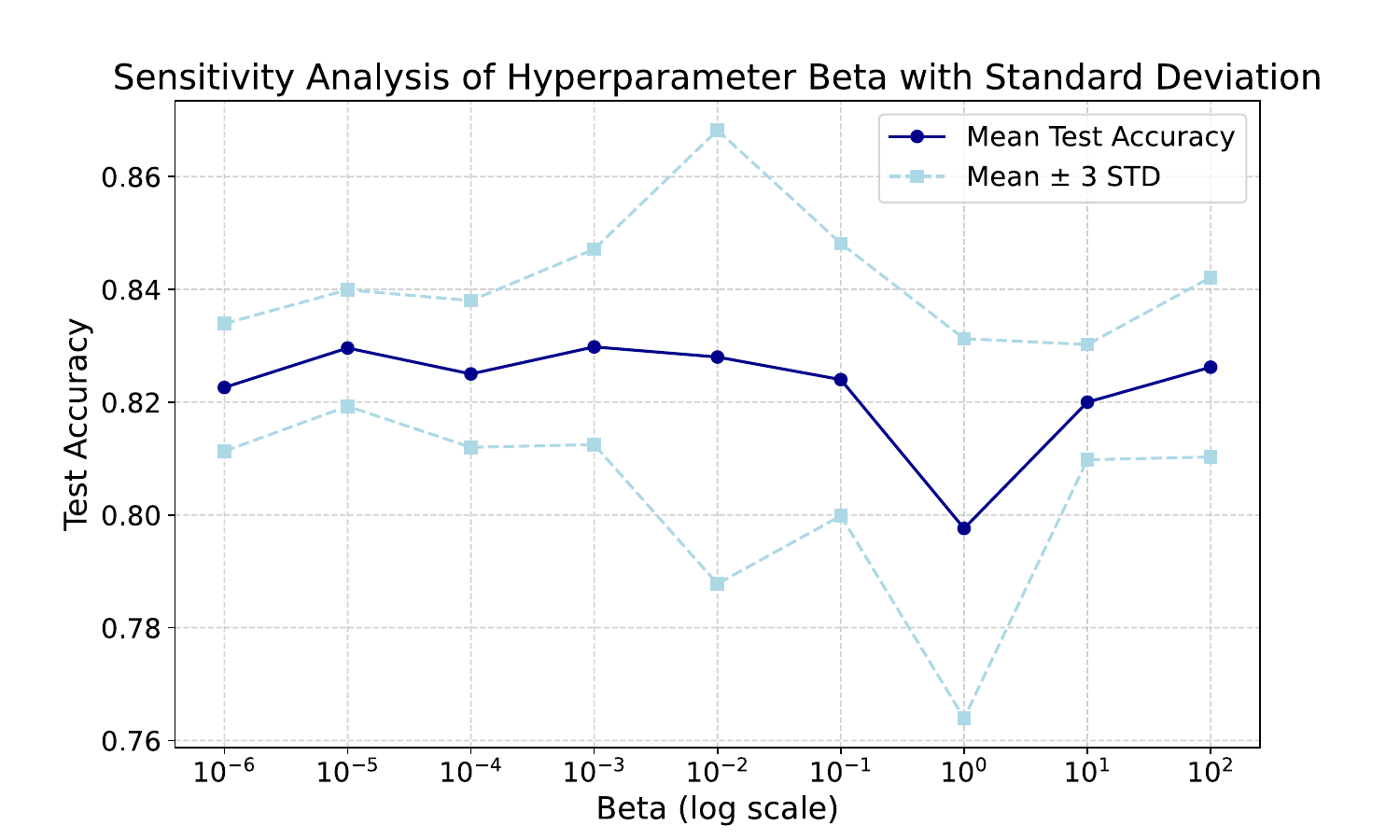} 
\caption{Sensitivity analysis of hyperparameter beta with standard deviation.}
\label{fig:sensitivity}
\end{figure}

To investigate the impact of the regularization constraint on our method, experiments were conducted on the Cora dataset. The influence of regularization within the loss function was controlled by varying the hyperparameter Beta, which ranged from \(10^{-6}\) to \(10^{2}\). The results, as illustrated in Figure \ref{fig:sensitivity}, indicate significant sensitivity to changes in Beta. Specifically, it was observed that values of Beta greater than or equal to \(10^{-5}\) had a pronounced effect on model performance. Optimal results on the Cora dataset were achieved when Beta was set within the magnitude of \(10^{-3}\). This analysis highlights that Beta critically balances the strength of regularization and the intensity of contrastive learning, ensuring an optimal trade-off that enhances model performance without compromising generalization capabilities.
 
\subsection{Sensitivity Analysis of the Subgraph Size}
\begin{figure}
\centering
\includegraphics[width=0.7\textwidth]{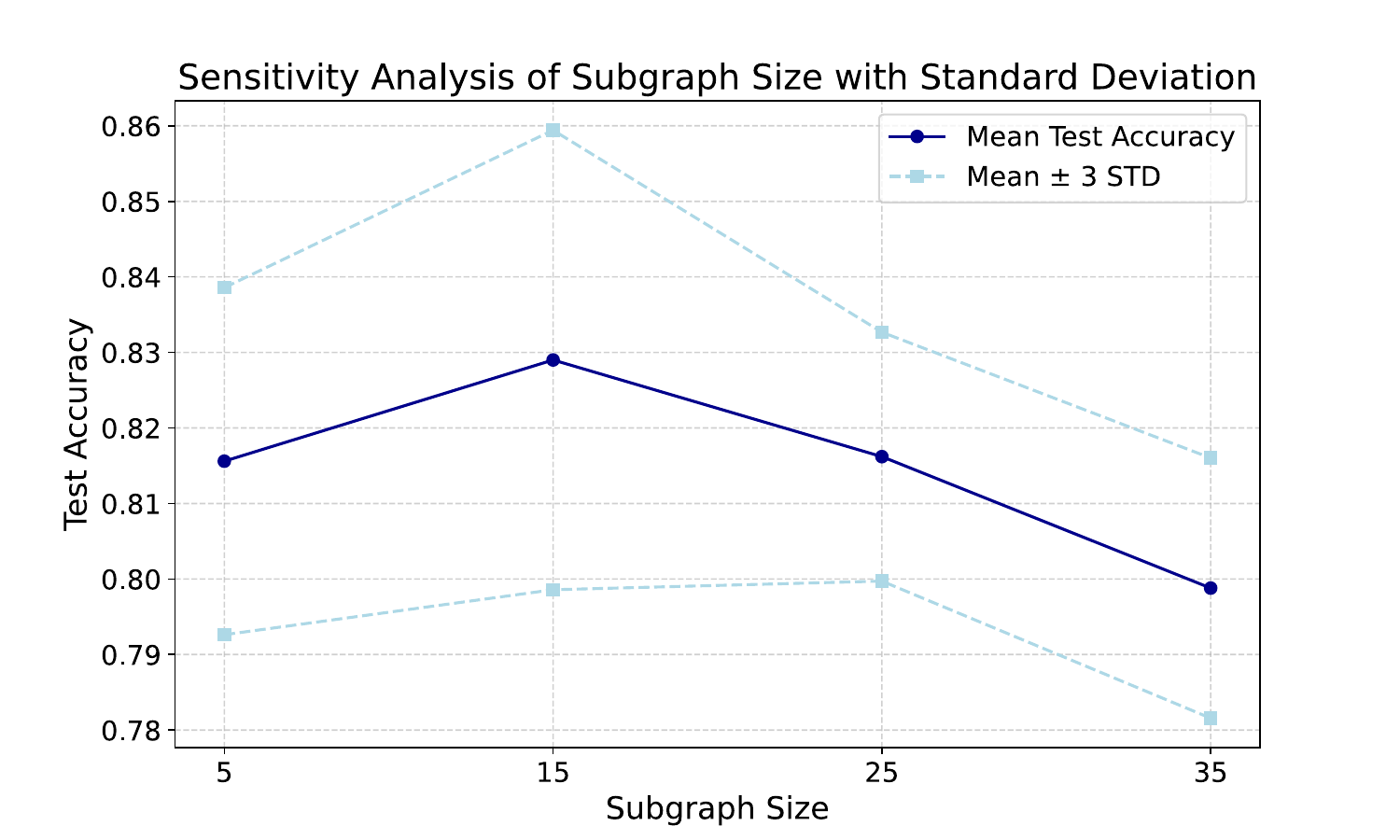}
\caption{Sensitivity analysis of subgraph size \( k \) on the Cora dataset. The plot shows the mean test accuracy (dark blue line) with error bars representing the mean \(\pm\) 3 standard deviations (light blue dashed lines) for different subgraph sizes.}
\label{fig:sensitivity_subgraph_size}
\end{figure}

To assess the impact of subgraph size on the performance of our \method~model, we conducted a sensitivity analysis using the Cora dataset with subgraph sizes \( k = 5, 15, 25, \) and \( 35 \).  These results, summarized in Figure~\ref{fig:sensitivity_subgraph_size}, indicate that a moderate subgraph size (\( k = 15 \)) yields the highest accuracy and lowest variability, suggesting that it effectively captures essential structural information without introducing excessive noise. Larger subgraphs might include redundant or irrelevant information, leading to slight decreases in performance. Therefore, selecting an appropriate subgraph size is crucial for optimizing the representation learning capability of \method.

\end{document}